\documentclass[sigconf]{acmart}

\AtBeginDocument{%
  \providecommand\BibTeX{{%
    Bib\TeX}}}

\usepackage{tikz}
\usepackage{makecell}
\usepackage{algorithm} 
\usepackage{microtype}
\usepackage{graphicx}
\usepackage{booktabs} 
\usepackage{algorithmic} 
\usepackage{tabularx}
\usepackage{array}  
\usepackage{booktabs}

\usepackage{colortbl}
\usepackage{textcomp}
\usepackage{graphicx}
\usepackage{textcomp}
\usepackage{xcolor}
\newcommand*\circlednum[1]{\tikz[baseline=(char.base)]{
    \node[shape=circle, fill=black, text=white, inner sep=1 pt, minimum size=1.1em] (char) { #1};}
}
\usepackage{ragged2e}

\usepackage{array}
\usepackage{comment}

\usepackage{graphicx}
\usepackage{textcomp}

\usepackage{subcaption}
\usepackage{xspace}
\usepackage{listings}
\usepackage{booktabs}
\usepackage{tabularx}
\usepackage{multirow}
\usepackage{booktabs}
\usepackage{colortbl}
\usepackage{tikz}

\usepackage{ctable}
\usepackage{listings}

\definecolor{MidnightBlue}{RGB}{25,25,112}

\makeatother
\usepackage[skip=2pt,font=small]{caption}
\usepackage{balance} 
\usepackage{nicematrix}
\usepackage{textcomp}
\usepackage{pifont}
\usepackage{float}
\usepackage{stfloats}
\usepackage{pifont}

\newcommand{\squishlist}{
   \begin{list}{$\bullet$}{%
        \setlength{\itemsep}{0pt}%
        \setlength{\parsep}{0pt}%
        \setlength{\topsep}{0pt}%
        \setlength{\partopsep}{0pt}%
        \setlength{\listparindent}{-2pt}%
        \setlength{\itemindent}{-5pt}%
        \setlength{\leftmargin}{1.2em}%
        \setlength{\labelwidth}{0em}%
        \setlength{\labelsep}{0.5em}%
    }
}
\usepackage{seqsplit}
\newcommand{\squishend}{
    \end{list}  }

\usepackage{seqsplit}

\usepackage{booktabs,array}
\newcolumntype{\ttP}[1]{>{\ttfamily\small\arraybackslash}p{#1}}

\newcommand {\nickname}{\textsc{PRO-V-R1}\xspace}

\setcopyright{none}
\def\BibTeX{{\rm B\kern-.05em{\sc i\kern-.025em b}\kern-.08em
    T\kern-.1667em\lower.7ex\hbox{E}\kern-.125emX}}
\usepackage{fontawesome5} 
\usepackage{hyperref}     
\usepackage{xcolor}       

\definecolor{linkblue}{HTML}{709CB4} 

\hypersetup{
    colorlinks=true,
    urlcolor=linkblue, 
    linkcolor=black,
    citecolor=black
}

\begin{document}

\title{\nickname:  \underline{R}easoning Enhanced \underline{Pro}gramming Agent for RTL \underline{V}erification }


\author{%
  Yujie Zhao$^{1}$,
  Zhijing Wu$^{1}$,
    Boqin Yuan$^{1}$,
 Zhongming Yu$^{1}$,
  Hejia Zhang$^{1}$,
  Wentao Ni$^{1}$,\\
  Chia\mbox{-}Tung Ho$^{2}$,
  Haoxing Ren$^{2}$,
  Jishen Zhao$^{1}$
}
\affiliation{%
  \institution{$^{1}$University of California San Diego \quad $^{2}$NVIDIA}
  \city{}
  \state{}
  \country{}
}
\email{{yuz285,b4yuan,zhy025,hez024, w2ni,jzhao}@ucsd.edu}
\email{neverakwu@gmail.com, chiatungh@nvidia.com, haoxingr@nvidia.com}

\begin{abstract}
Register-Transfer Level (RTL) verification is a primary bottleneck consuming 60-70\% of development time. While Large Language Models (LLMs) show promise for RTL automation, their performance and research focus have overwhelmingly centered on RTL generation rather than verification.  Current methods for RTL verification rely on large scale proprietary models (e.g., GPT-4o) to generate Python-based functional references, incurring a high cost and
raising data-privacy risks. To date, an end-to-end open-source solution for autonomous verification remains absent.

We introduce \nickname, the first trainable open-source agentic framework for autonomous RTL verification. Our contributions are threefold: (1) we design PRO-V~sys, a modular agentic system that couples LLM-based reasoning with programmatic tool use for RTL verification; (2) we establish a data construction pipeline that leverages existing RTL datasets to build simulation-validated, expert-level trajectories tailored for supervised fine-tuning (SFT) RTL verification agents; and (3) we implement an efficient reinforcement learning (RL) algorithm that uses verification-specific rewards derived from program-tool feedback to optimize the end-to-end verification workflow. Our empirical evaluation demonstrates \nickname achieves a \textbf{57.7\%} functional correctness rate and \textbf{34.0\%} in robust fault detection, significantly outperforming the base model's 25.7\% and 21.8\% (respectively) from the state-of-the-art (SOTA) automatic verification system. This configuration also outperforms large-scale proprietary LLMs in functional correctness and shows comparable robustness for fault detection.
\end{abstract}
\maketitle
\noindent
\faGithub\ \textbf{Pro-V Training Code:} \\
\url{https://github.com/pettingllms-ai/PettingLLMs}

\vspace{0.3em} 

\noindent
\faGithub\ \textbf{Pro-V Agentsys Code:} \\
\url{https://github.com/stable-lab/Pro-V}
\section{introduction}\label{sec:intro}

Hardware design verification is a critical stage in the VLSI design flow, ensuring that a Register-Transfer Level (RTL) implementation faithfully conforms to its functional specification. In simulation-based verification, this process fundamentally relies on an executable testbench that applies constrained stimuli to the Design Under Test (DUT), monitors its responses, and compares them against the expected behavior; the quality of this testbench largely determines which functional defects can be exposed before tape-out~\cite{bergeron2003writing,ioannides2012coverage}. However, constructing effective testbenches remains a labor-intensive process, which often accounts for 60\% to 70\% of the total development time, representing a significant bottleneck in the modern hardware design cycle~\cite{shin2020efficient}.


\begin{figure}[t]
    \centering
    
    \includegraphics[width=1\linewidth]{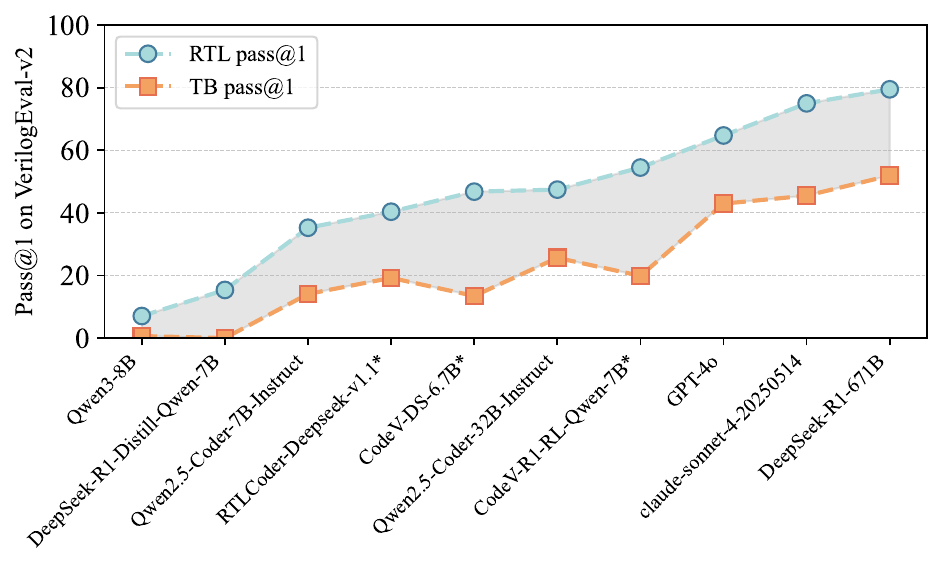}
  \caption{Performance comparison of different LLMs on RTL (green line) and testbench (blue line) generation. Models marked with * are RTL-specific. TB pass@1 denotes that the golden reference RTL code passes the LLM-generated testbench.}
\label{fig:model_performance_comparison}

\end{figure}
Recent advances in large language models (LLMs) for verilog enable a promising path to automating verification workflows. Nevertheless, the vast majority of current research targets RTL generation, yielding numerous benchmark datasets~\cite{liu2023verilogeval,rtllm,verilogeval-v2-arxiv} and specialized models~\cite{thakur2024verigen, liu2023rtlcoder,codev-r1,zhao2024codev,zhu2025codevr1}.In parallel with this distribution of research effort, we observe a divergence in model capabilities. As depicted in Fig.~\ref{fig:model_performance_comparison}, contemporary LLMs perform consistently better in RTL generation (green line) compared to testbench generation (blue line).

While testbench generation is critical and presents a significantly greater challenge, research into LLM-based verification still heavily relies on commercial LLMs. To navigate the difficulty of direct hardware description generation, current efforts are converging on a more pragmatic paradigm: instead of generating complete Verilog testbenches, these approaches use LLMs to generate a \textit{Python-based functional reference model (FRM)} for the DUT\cite{qiu2024autobench,qiu2024correctbench,tan2025autoverifix}, which emulates its intended functional behavior. This Python-centric design is capability-driven, as modern code LLMs are substantially stronger on Python than on Verilog~\cite{chen2021humaneval, liu2023verilogeval}.

\begin{figure}[t]
    \centering
    
    \includegraphics[width=1\linewidth]{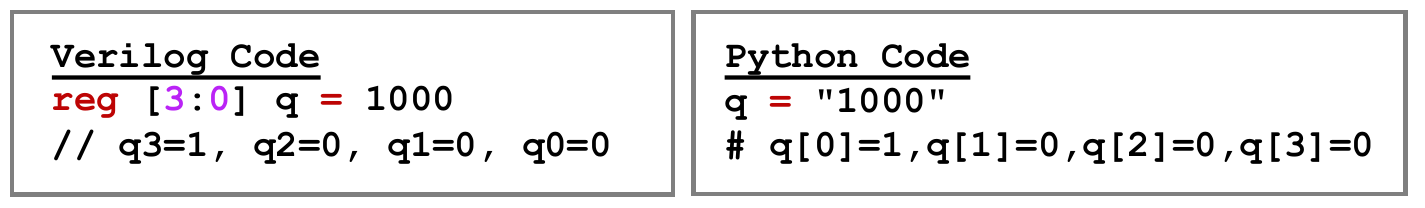}
   \caption{Semantic gap in indexing and slicing: Verilog's significance-based \texttt{q[3]} (MSB) contrasts with Python's offset-based \texttt{q[0]}. Also, Verilog's inclusive-range slice \texttt{q[3:2]} (`10') maps to Python's half-open range \texttt{q[0:2]}.}
    \label{fig: bit}
    \vspace{-10 pt}

    \label{fig:enter-label}
\end{figure}

However, this reliance on large-scale proprietary models (e.g., GPT-4o) introduces critical limitations. Beyond high inference costs and data-privacy risks for sensitive IPs, the inability to fine-tune these generalist ``black-box'' models leads to frequent hallucinations, particularly where Python and Verilog semantics diverge (e.g., conflicting indexing and slicing mechanics, as shown in Fig. \ref{fig: bit}).  While open-source efforts focus on RTL generation, verification lacks comparable open-source datasets and training frameworks. Consequently, a fully open-source, end-to-end solution for autonomous testbench generation remains absent.

A clear path forward is to train open-source models capable of mastering the autonomous RTL verification workflow. However, realizing this goal presents three significant challenges due to the complexity of hardware verification:  (i) Agentic System Inefficiency: SOTA baselines like CorrectBench~\cite{qiu2024correctbench} are computationally expensive (avg. 805 s/task) while using open-source model and are prone to storage explosion from complex compilation processes; (ii) Supervised Fine-tuning (SFT) Data Scarcity: Existing public datasets are predominantly tailored for RTL design and implementation, resulting in a critical lack of ground-truth verification trajectories required to effectively fine-tune agents for verification tasks; and (iii) Missing RL Training Framework: The field lacks a dedicated training framework to optimize agentic verification policies via reinforcement learning.

To address these challenges, we introduce \nickname: the first end-to-end open-source solution for autonomous RTL verification that enhances LLM reasoning with programmatic tools. Our contributions are as follows:
\squishlist \item We design PRO-V sys, a novel agentic system that enhances LLM reasoning with programmatic tool for autonomous RTL verification. 
\item We establish a data construction pipeline that leverages existing RTL datasets to construct verified expert-level trajectories tailored for the SFT of RTL verification agents.

\item We implement an efficient RL algorithm using verification-specific structured rewards from programmatic tool feedback to optimize the end-to-end verification workflow. \squishend

Empirically, \nickname achieves \textbf{57.7\%} functional correctness and \textbf{34.0\%} fault detection, outperforming SOTA agent system baselines (25.7\%/21.8\%) and matching or exceeding proprietary LLMs. Moreover, it provides an \textbf{8.8$\times$ speedup}, averaging just \textbf{91 s/task}.

\section{Background and Motivation}
\label{related-work}


\subsection{LLM-based Verilog generation and verification}
\label{sec:llm-based tb generation}
The preponderance of research in LLMs for RTL has concentrated on RTL code generation, including significant open-source efforts such as benchmark datasets~\cite{liu2023verilogeval,rtllm,verilogeval-v2-arxiv} and specialized models~\cite{thakur2024verigen, liu2023rtlcoder,codev-r1,zhao2024codev,zhu2025codevr1}.
In contrast, the emerging paradigm of LLM-aided RTL verification has shifted from direct HDL generation—a known challenge~\cite{chen2021humaneval, liu2023verilogeval}—towards generating Python-based FRMs~\cite{qiu2024autobench, qiu2024correctbench, tan2025autoverifix}  within agent-controllable verification environments.
However, this approach relies almost exclusively on large-scale, proprietary LLMs (e.g., GPT-4o)~\cite{qiu2024autobench, tan2025autoverifix}.
This dependency introduces high inference costs, data-privacy risks, and reliability issues; as ``black-box" generalists, they cannot be fine-tuned and are prone to hallucinations where Python and Verilog semantics diverge.
Consequently, agent systems, curated datasets, and training frameworks for LLM-aided RTL verification are scarce, leaving the community without open, reproducible models specialized for this new functional verification flow.
\subsection{Tool-Enhanced LLM Agents with Post-training}
\label{sec: relatedwork-toolrl}
Tool-enhanced agents leverage LLMs as controllers to interact with external, domain-specific tools~\cite{schick2023toolformer, mialon2023augmented}, often structured within a modular ``agent flow" (e.g., planner, executor) to solve complex tasks~\cite{li2025agentflow}. An agent's effectiveness hinges on clear tool APIs~\cite{qin2024toolllm, patil2023gorilla} and a robust policy for tool invocation. This policy is typically established via SFT on expert demonstrations, and then refined using RL from environmental feedback~\cite{lu2023adapt, zeng2024agentic}. The RL phase optimizes the policy by assigning a scalar "verified reward" to the agent's output, favoring actions that yield high-reward outcomes.

While this SFT-then-RL paradigm has been successfully applied in diverse domains, including mathematical reasoning~\cite{gao2022pal, chen2022pot}, web navigation~\cite{he2024webvoyager, zhou2023webarena}, and scientific discovery~\cite{bran2023chemcrow, lu2024aiscientist}, its application to hardware verification remains largely unexplored. This gap exists because a complete, effective training loop for verification requires addressing three critical, domain-specific challenges: (i) overcoming the scarcity of high-quality, verified reasoning traces for SFT; (ii) designing a trainable agent loop with a reward function that translates sparse simulation outcomes into a dense learning signal; and (iii) developing an effective optimization algorithm for credit assignment within the complex, multi-step agentic workflow. Our work is the first to bridge this gap by establishing this complete loop.

\begin{figure*}[t]
    \centering
    \includegraphics[width=\linewidth]{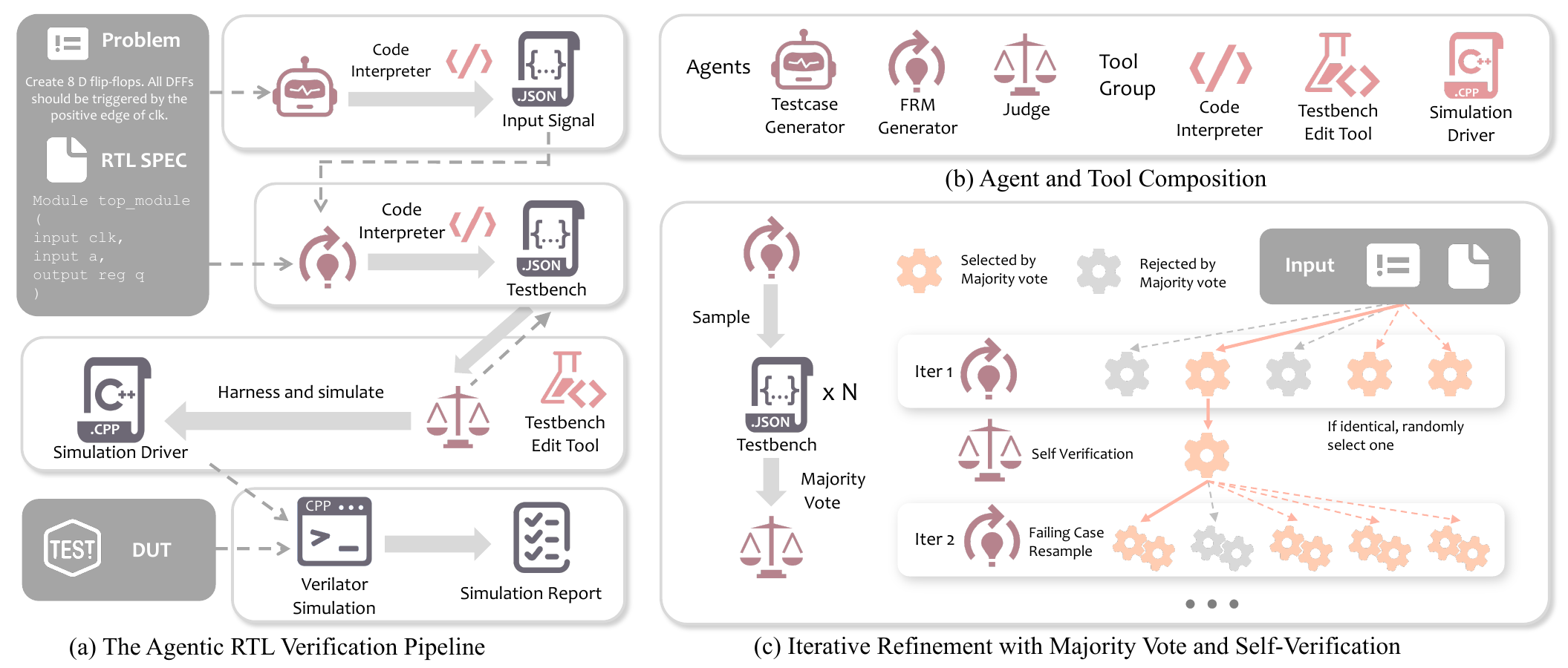}
    \caption{Overview of the autonomous agent-based framework for RTL design verification.
    \textbf{(a) The Agentic RTL Verification Pipeline}, detailing the end-to-end process from high-level prompt and RTL specification to a final simulation report.
    \textbf{(b) Agent and Tool Composition}, defining the set of agentic roles and their corresponding callable tools.
    \textbf{(c) Iterative Refinement with Majority Vote and Self-Verification}, illustrating the sampling, selection, and self-correction loop used to enhance the quality of generated testbenches.}
    \label{fig:framework}
\end{figure*}

\begin{table*}[t]
\centering
\caption{Summary of Program-Based Tool APIs.}
\label{tab:prove_tools}

\begin{tabular}{@{} l l p{0.24\textwidth} p{0.28\textwidth} p{0.13\textwidth} @{}}
\toprule
\textbf{Tool Name} & \textbf{Tool API} & \textbf{Argument Schema} & \textbf{Example Argument} & \textbf{Output Artifact} \\ \midrule

Code Interpreter & \texttt{\textless python\textgreater} &
Python code snippet &

\texttt{"def step(x): return ((x<<1) \string^ 0x3D) \& 0xFF"} &
Execution Output \\ \midrule

Testbench Editor & \texttt{\textless tb\_edit\textgreater} &

\texttt{op} (add\textbar replace\textbar remove); \texttt{target} (JSON Pointer of the field to modify); \texttt{value} (new content for add/replace). &
\texttt{"op":"add", "target":"error location",\allowbreak "value": \allowbreak\{"inputs":\{"y":"001","w":"0"\},\allowbreak "expected\_outputs":\{"Y2":"0"\}\}} &
Updated Testbench  \\ 
\bottomrule
\end{tabular}
\end{table*}

\section{Methods}
\label{sec:design}

To address the challenges in Sec.~\ref{sec:intro} and ~\ref{related-work}, we introduce \nickname, a program-tool-enhanced agentic framework for autonomous RTL verification, composed of two primary components: (1) an agentic verification system that instantiates modular agents equipped with specialized programmatic tools to interface with industrial simulators; and (2) an agentic training framework for RTL verification to optimize the full agentic workflow.

\subsection{Agentic Verification System}
\label{sec:pipeline}
Our framework executes RTL verification through an autonomous agentic system, as illustrated in Fig~\ref{fig:framework}. Leveraging the LLMs' tool-use capabilities and the industry-standard simulators like Verilator, our system mitigates the inherent limitations and hallucination associated with LLMs directly generating complex hybrid RTL testbenches, thereby enhancing the fidelity and correctness of the generated verification artifacts. This process integrates a society of agents with a specialized toolset (Fig.~\ref{fig:framework}~(b)). The architecture is composed of three distinct agentic roles: the \underline{Testcase Generator}, the \underline{FRM Generator}, and the \underline{Judge}. These agents are equipped with a Tool Group comprising the \textit{Code Interpreter}, the \textit{Testbench Edit Tool}, and the \textit{Simulation Driver}. The end-to-end verification flow (Fig.~\ref{fig:framework}~(a)), detailed below, is executed in three main stages:

\noindent\textbf{\circlednum{1} Testcase Generation.} LLMs exhibit known weaknesses in operations requiring high numerical fidelity, such as precisely generating long numerical sequences (e.g., 256-bit vectors) ~\cite{stringllm2024,xu2025count}. This inability directly conflicts with the stringent demands of RTL verification, which requires bit-accurate, width-sensitive (e.g., 128/512-bit), and multi-cycle stimuli. To overcome this challenge, the \underline{Testcase Generator} (Fig.~\ref{fig:framework}~(b)) adopts a program generation approach. As shown in the first stage of Fig.~\ref{fig:framework}~(a), instead of attempting to directly emit long bit strings, the agent first reasons about the desired test scenario. It then executes a short Python script via the \textit{Code Interpreter} to programmatically construct the required input sequences. This approach guarantees bit-accuracy, bypasses the LLM's fidelity limitations, and produces an \texttt{Input Signal} with functionally valid, coverage-driven test vectors.

\noindent\textbf{\circlednum{2} Testbench Generation and Functional Modeling.} The \underline{FRM Generator} (Fig.~\ref{fig:framework}~(b)) takes the \texttt{JSON Input Signal} from step~\circlednum{1} and the \texttt{RTL SPEC} to construct a \texttt{testbench.json} artifact, pairing input vectors with cycle-aligned expected outputs. To achieve this, the agent leverages LLM-based code generation (Sec.~\ref{sec: relatedwork-toolrl}) to sample $N$ Python-based FRMs acting as ``golden'' behavioral oracles. These scripts are executed by the \textit{Code Interpreter} to compute $N$ sets of expected outputs for the step~\circlednum{1} testcases. This program-aided approach circumvents LLM limitations in direct arithmetic~\cite{gao2023pal,li2024gsmplus} by applying precise, executable logic.

\noindent\textbf{\circlednum{3} Iterative Refinement and Verification.} The \underline{Judge} agent (Fig.~\ref{fig:framework}~(b)) receives the $N$ candidate artifacts from step~\circlednum{2} and drives a self-improvement loop (Fig.~\ref{fig:framework}~(c)) combining majority voting with verification. First, the system selects the most consistent candidate via majority voting. The \underline{Judge} then validates the selected outputs; if errors are detected, it performs targeted corrections by invoking the \textit{Testbench Edit Tool}. As detailed in Tab.~\ref{tab:prove_tools}, this tool allows the agent to programmatically \texttt{add}, \texttt{replace}, or \texttt{remove} entries in \texttt{testbench.json} (e.g., rectifying a single failing vector). Upon validation or budget exhaustion, the final \texttt{testbench.json} is passed to the \textit{Simulation Driver} to compile and orchestrate the Verilator simulation. This driver validates the DUT by comparing its actual execution results—driven by the stimuli from step~\circlednum{1}—against the expected reference outputs encapsulated in the \texttt{testbench.json}.

\subsection{Agentic training framework for RTL verification}
\label{sec:training}
While the agentic system detailed in Sec.~\ref{sec:pipeline} establishes a robust verification loop, its effectiveness is inherently constrained by the reliance on general-purpose LLMs. These models struggle with precise instruction following (e.g., tool arguments, role confusion), lack specialized RTL domain knowledge (e.g., bit-order conventions), and possess weak intrinsic logical and coding abilities for model generation and debugging. To evolve these generalists into domain experts, we propose a hybrid training methodology: (i) initiating with SFT on verified, expert-level trajectories synthesized from existing RTL datasets via a custom data construction pipeline, and (ii) implementing an efficient RL algorithm that leverages verification-specific structured rewards derived from programmatic tool feedback to optimize the end-to-end workflow.
\begin{figure}[t]
    \centering
    \includegraphics[width=\linewidth]{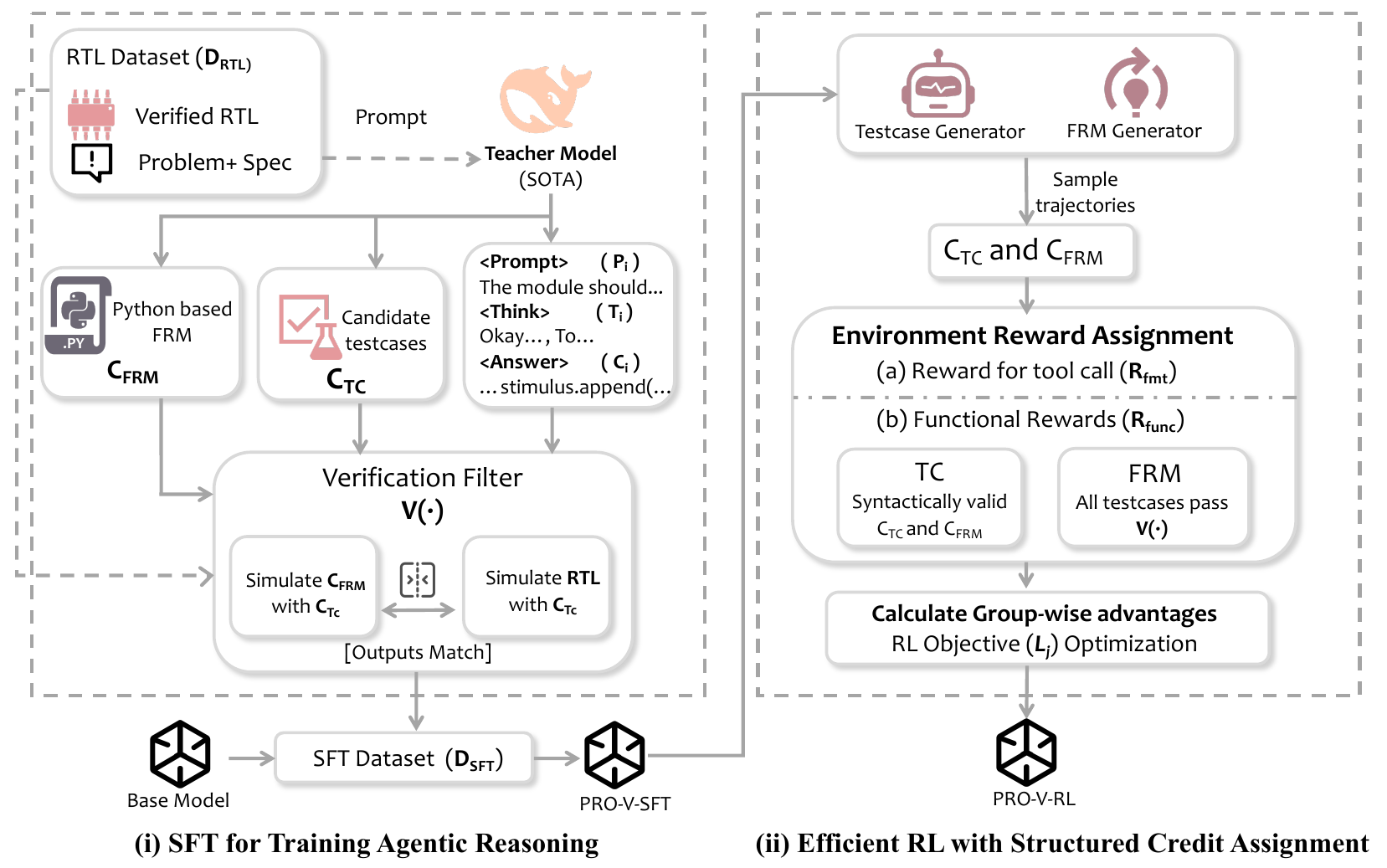}
    \caption{Overview of Agentic training framework.}
    \label{fig:framework_train}
 
\end{figure}

\textbf{(i) SFT for Agentic Reasoning.}
Existing public datasets for hardware design, such as those focused on Verilog generation~\cite{zhu2025qimengcodev, craft-rl} (as illustrated in Sec.~\ref {sec:llm-based tb generation}), primarily consist of verified (Problem, RTL) pairs, which we denote as $\mathcal{D}_{RTL}$. While suitable for simple generation tasks, these datasets lack the comprehensive reasoning traces and hardware verification references required to train our specific agentic workflow.

To construct our SFT dataset, we leverage these existing datasets as a foundation and establish a dedicated data construction pipeline to synthesize the verification specific expert-level reasoning trajectories and tool interactions. For each problem, we use its golden RTL design and prompt a SOTA `teacher' model to generate trajectories for two agent roles: the \underline{Testcase Generator} and the \underline{FRM Generator}. The teacher model generates candidate testcases ($C_{TC}$) and a corresponding Python-based FRM, $C_{FRM}$. A trace is only accepted after passing a strict \underline{verification filter}, $\mathcal{V}(\cdot)$, which is defined as:
\begin{align}
\label{eq:simulate}
   \nonumber &\mathcal{V}(C_{TC}, C_{FRM}, RTL)\\
        \equiv & [\text{Sim}(RTL, C_{TC}) = \text{Sim}(C_{FRM}, C_{TC})]
\end{align}
This filter ensures the agentic flow successfully generates a runnable test harness (composed of $C_{TC}$ and $C_{FRM}$) and that the golden RTL's simulation output precisely matches the FRM's output for all testcases $C_{TC}$. This process validates the functional correctness of the generated $C_{FRM}$. We then collect these verified interactions to form our curated SFT dataset, $\mathcal{D}_{SFT}$. Each data point $d \in \mathcal{D}_{SFT}$ is a trajectory composed of components for each agent, structured as: $
(P_i, T_i, C_i), \text{where} \,i \in \{\text{TC}, \text{FRM}\}$.
Here, $P_i$ represents the prompt, $T_i$ is the reasoning trace, and $C_i$ is the generated code for agent $i$.

The SFT objective trains the model parameters $\theta$ by minimizing the standard log-likelihood loss over all prompt-response pairs $(x, y)$ derived from $\mathcal{D}_{SFT}$ (e.g., $x=P_i$, $y=T_i \oplus C_i$, where $\oplus$ denotes concatenation):
\begin{align}
    \mathcal{L}_{SFT}(\theta) = - \sum_{(x, y) \in \mathcal{D}_{SFT}} \sum_{t=1}^{|y|} \log P_\theta(y_t | y_{<t}, x)
\end{align}
where $y_t$ is the $t$-th token of the target response $y$.

\textbf{(ii) Efficient RL with Structured Credit Assignment.}
While SFT provides the model with a foundational capability to mimic expert trajectories in $\mathcal{D}_{SFT}$, we introduce a subsequent reinforcement learning phase to enable our agents to generalize and discover policies that maximize verification success. We utilize Group Relative Policy Optimization (GRPO)~\cite{shao2024deepseekmath,deepseekr1} to fine-tune the SFT model through direct interaction with the verification environment. 

For a given task $q$, we sample $G$ complete trajectories $\{\tau_i\}_{i=1}^G$. Each trajectory $\tau_i$ is a composition of the sub-trajectories from its constituent agents: $\tau_i = \tau_{i, \text{TC}} \oplus \tau_{i, \text{FRM}}$, representing the full generation from the \underline{Testcase Generator} and the \underline{FRM Generator}.

To enable precise credit assignment, we define distinct rewards for each agent $j \in \{\text{TC}, \text{FRM}\}$. The reward for each agent's sub-trajectory $\tau_{i,j}$ is a composite signal:
\begin{equation}
\label{eq:reward-composite}
R(\tau_{i,j}) = R_{func}^{j}(\tau_i) + w_{fmt} R_{fmt}^{j}(\tau_{i,j})
\end{equation}
where $\tau_i = \tau_{i, \text{TC}} \oplus \tau_{i, \text{FRM}}$ is the full trajectory, $R_{fmt}^{j}(\tau_{i,j})$ is the agent-specific reward for generating syntactically correct tool calls, and $w_{fmt}$ is a weighting hyperparameter. The functional rewards $R_{func}^{j}$ are computed separately for each agent based on the full trajectory's outputs ($C_{TC}$ and $C_{FRM}$):
\begin{itemize}
    \item \textbf{TC Agent Functional Reward ($R_{func}^{\text{TC}}$):} This is a binary reward for generating a \textit{runnable harness}. It is $1$ if $C_{TC}$ and $C_{FRM}$ are syntactically valid and can be successfully compiled and simulated together, and $0$ otherwise.
    \item \textbf{FRM Agent Functional Reward ($R_{func}^{\text{FRM}}$):} This is the primary verification success reward. It is $1$ only if \textit{all test cases pass} verification—that is, the strict filter $\mathcal{V}(\cdot)$ (defined in Eq.~\eqref{eq:simulate}) evaluates to true. This reward is contingent on the harness being runnable (i.e., $R_{func}^{\text{TC}}=1$).
\end{itemize}

For each agent $j\in\{\mathrm{TC},\mathrm{FRM}\}$, let $\{\tau_{i,j}\}_{i=1}^{G}$ be a minibatch of sub-trajectories and $R(\tau_{i,j})$ the composite reward in Eq.~\eqref{eq:reward-composite}. We normalize rewards \emph{per agent}:
\begin{equation}
\label{eq:grpo-adv}
A_{i,j}
=\frac{R(\tau_{i,j})-\bar{R}_j}{\max\!\big(\mathrm{std}(\{R(\tau_{k,j})\}_{k=1}^{G}),\,\varepsilon\big)},
\end{equation}
where $\bar{R}_j$ and $\mathrm{std}(\cdot)$ are the mean and standard deviation over agent $j$'s batch, and $\varepsilon>0$ avoids division by zero. 

The GRPO objective sums agent-wise clipped-ratio losses:
\begin{align}
\label{eq:grpo-loss}
\mathcal{L}_{\mathrm{GRPO}}(\theta)
  &= \sum_{j \in \{\text{TC}, \text{FRM}\}} \mathcal{L}_j(\theta) \\
\text{where}\quad 
\mathcal{L}_j(\theta)
  &= \mathbb{E}_{i,t \in \tau_{i,j}}
     \biggl[
       \min\biggl(
         \rho_{i,t} A_{i,j},\;
\\[-6pt] &\qquad\qquad
         \mathrm{clip}(\rho_{i,t}, 1-\lambda, 1+\lambda) A_{i,j}
       \biggr)
     \biggr]
     \notag
\end{align}

with per-token likelihood ratio
$\rho_{i,t}=\frac{\pi_{\theta}(a_{i,t}\mid s_{i,t})}{\pi_{\theta_{\mathrm{old}}}(a_{i,t}\mid s_{i,t})}$
and clipping parameter $\lambda>0$. This critic-free formulation yields stable updates aligned with the agent-specific rewards.

\section{Experiments}
\subsection{Experimental Setup}
\label{sec:exp-setup}
\begin{table*}[b]
\centering
\caption{Overall results on VerilogEval-v2 and RTLLM v2.0.}
\label{tab:key-results}

\scriptsize
\resizebox{\textwidth}{!}{%
\begin{tabular}{llcrrrrrrrrrr}
\toprule
\multirow{2}{*}{Category} &
\multirow{2}{*}{Model} &
\multirow{2}{*}{\shortstack{Open\\Source}} &
\multicolumn{5}{c}{VerilogEval-v2} &
\multicolumn{5}{c}{RTLLM v2.0} \\
\cmidrule(lr){4-8}\cmidrule(lr){9-13}
& & & Eval0 & Eval1 & Eval2-80 & Eval2-90 & Eval2-100 & Eval0 & Eval1 & Eval2-80 & Eval2-90 & Eval2-100 \\
\midrule

\multirow{7}{*}{Foundation models}
& GPT-4o & $\times$ 
  & 90.4 & 43.0 & 39.7 & 36.5 & 27.6
  & \textbf{90.0} & \textbf{42.0} & \textbf{18.0} & \textbf{14.0} & \textbf{6.0} \\

& DeepSeek-R1-671B* & $\checkmark$
  & 87.2 & \underline{51.9} & \textbf{44.9} & \textbf{40.4} & \textbf{35.9}
  & 72.0 & 28.0 & \underline{16.0} & 10.0 & \textbf{6.0} \\

& claude-sonnet-4-20250514 & $\times$
  & \textbf{94.9} & 45.5 & 41.7 & \underline{39.7} & \textbf{35.9}
  & \underline{88.0} & \underline{30.0} & 10.0 & 4.0 & 2.0 \\

& DeepSeek-R1-Distill-Qwen-7B* & $\checkmark$
  & 0.0 & 0.0 & 0.0 & 0.0 & 0.0
  & 0.0 & 0.0 & 0.0 & 0.0 & 0.0 \\

& Qwen3-8B & $\checkmark$
  & 0.6 & 0.6 & 0.6 & 0.6 & 0.0
  & 0.0 & 0.0 & 0.0 & 0.0 & 0.0 \\

& Qwen2.5-Coder-7B-Instruct & $\checkmark$
  & 70.5 & 14.1 & 10.3 & 9.6 & 7.7
  & 58.0 & 8.0 & 0.0 & 0.0 & 0.0 \\

& Qwen2.5-Coder-32B-Instruct & $\checkmark$
  & 72.4 & 25.6 & 22.4 & 19.9 & 18.0
  & 66.0 & 8.0 & 4.0 & 2.0 & 2.0 \\

\midrule
\multirow{3}{*}{Domain specific models}

& RTLCoder-Deepseek-v1.1 & $\checkmark$
  & 75.0 & 19.2 & 12.8 & 10.9 & 8.3
  & 70.0 & 6.0 & 0.0 & 0.0 & 0.0 \\

& CodeV-R1-RL-Qwen-7B & $\checkmark$
  & 81.4 & 19.9 & 15.4 & 11.5 & 10.3
  & 84.0 & 4.0 & 2.0 & 2.0 & 2.0 \\

& CodeV-DS-6.7B & $\checkmark$
  & 37.2 & 13.5 & 6.4 & 3.2 & 2.6
  & 64.0 & 20.0 & 4.0 & 0.0 & 0.0 \\

\midrule
\multirow{2}{*}{Ours}

& PRO-V-DS-8B w/ PRO-V sys & $\checkmark$
  & 93.6 & 50.6 & 35.9 & 34.0 & 27.6
  & 78.0 & 26.0 & \underline{16.0} & \underline{12.0} & \underline{4.0} \\

& PRO-V-R1-8B w/ PRO-V sys & $\checkmark$
  & \underline{94.8} & \textbf{57.7} & \underline{44.2} & 37.8 & \underline{34.0}
  & 78.0 & 28.0 & \textbf{18.0} & \textbf{14.0} & \textbf{6.0} \\

\bottomrule
\end{tabular}
}

\vspace{3pt}
\begin{minipage}{0.98\textwidth}
\footnotesize
\textbf{Bold} numbers indicate the best result in each column, and 
\underline{underlined} numbers indicate the second-best.
Rows not annotated with ``w/ PRO-V sys'' report vanilla-model performance.
``PRO-V-DS-8B'' denotes the SFT-only stage, while ``PRO-V-R1-8B'' denotes the subsequent RL-tuned model.

\end{minipage}
\vspace{-10pt}
\end{table*}
\noindent\textbf{Hardware.}
All training and evaluation are conducted on NVIDIA H100 GPUs in a single-node setup.  \textbf{Simulator.}
We use Verilator 5.042 (2025-11-02)~\cite{verilator_citation} as the cycle-accurate simulator. \textbf{Models.}
Our \emph{teacher} model for SFT data generation is \textsc{DeepSeek-V3.2} (Oct.\ 2025)~\cite{deepseekv3_2025}, accessed via its official API. The \emph{base} model is \textsc{Qwen3-8B}~\cite{qwen3_2025}.  \noindent\textbf{SFT Protocol.}
We train \textsc{Qwen3-8B}\cite{qwen3-8b} using the AdamW optimizer (learning rate $2\times10^{-5}$). We use a global batch size of 256, bfloat16 mixed precision, and gradient clipping at 1.0. \textbf{RL (GRPO) Protocol.} Key hyperparameters include: FRM sample number $N=5$, learning rate $1\times10^{-6}$, clipping parameter $\lambda=0.2$, group size $G=4$, format reward weight $w_{fmt}=0.3$, and $\varepsilon=1\times10^{-8}$ for advantage normalization. The maximum prompt and response lengths are both 4k tokens. During sampling for training rollouts, we use temperature $1.0$ and top-$p=0.95$; for evaluation, we use temperature $0.0$. \noindent\textbf{Benchmarks.}
We employ two primary benchmarks: VerilogEval-v2~\cite{verilogeval-v2-html} and RTLLM v2.0~\cite{rtllm}. To rigorously evaluate \textbf{testbench robustness}—a key metric for verification—we utilize DeepSeek-V3.2 to generate a set of 10 RTL mutants per problem for both datasets. The mutants are essential for assessing a testbench's ability to detect faults, moving beyond simple compilation. \noindent\textbf{Metrics.}
Following the methodology of CorrectBench~\cite{qiu2024correctbench}, we adopt its three-level protocol on VerilogEval-v2 to assess distinct aspects of \textbf{testbench quality}:
\textsc{Eval0} (Viability): Assesses fundamental syntactical and simulation viability. It reports the fraction of TBs that successfully compile and execute without simulation runtime errors. This serves as a baseline check for structural correctness.
\textsc{Eval1} (Functional Correctness): Assesses the primary correctness of the \emph{generated} testbench. It reports the fraction of TBs that successfully compile and pass simulation when run against the golden RTL code in the benchmarks. This metric validates the TB's ability to correctly "pass" a known-good design.
\textsc{Eval2-80/90/100} (Robustness \& Fault Detection): Quantifies the testbench's fault-detection capability using the generated RTL mutants. This metric reports the fraction of problems where the generated TB's verdict (pass/fail) matches the reference TB's verdict on at least 80\%, 90\%, and 100\% of the mutants, respectively. This metric, therefore, evaluates the testbench's critical ability to identify faulty RTL implementations.

\subsection{Key Results}

Tab.~\ref{tab:key-results} shows that \nickname with PRO-V system consistently matches or exceeds much larger proprietary models on end-to-end testbench quality. On VerilogEval-v2, PRO-V-R1-8B attains $57.7\%$ on Eval1 and $34.0\%$ on Eval2-100, outperforming GPT-4o and all existing open models, and approaching the best results of multi-hundred-billion–parameter systems such as DeepSeek-R1 and Claude. More importantly, our method maintains relatively strong robustness: the gap between Eval1 and Eval2 metrics is significantly smaller than for most baselines, especially on RTLLM v2.0, where GPT-4o and other foundations degrade sharply from high Eval1 to very low Eval2 scores, indicating that they often emit short, low-diversity stimuli that can validate the golden RTL but rarely expose mutants. In contrast, the PRO-V agents learn to construct richer Python testbenches that both compile reliably (high Eval0) and exercise corner cases (higher Eval2-80/90/100), while domain-specific baselines such as RTLCoder and CodeV either underperform on Eval1 or collapse on Eval2, reflecting insufficient fault coverage. A remaining limitation is that on the hardest VerilogEval-v2 robustness setting (Eval2-100), very large proprietary models still retain a small advantage, suggesting headroom for further scaling of training data and RL optimization, but \nickname already narrows the verification gap with a compact, fully open 8B model.

\subsection{Ablation Study}
\begin{table}[t]
  \centering
  \small
  \caption{Ablation Study of the PRO-V Framework on VerilogEval-v2. We progressively add our agentflow (PRO-V Flow), (SFT), and RL to the base model.}
  \label{tab:ablation_prov_framework}

  \begin{tabularx}{\columnwidth}{>{\RaggedRight}Xrrrrr}
    \toprule
    & \multicolumn{5}{c}{\textbf{VerilogEval-v2}} \\
    \cmidrule(lr){2-6}
    \textbf{Model / Method} & \textbf{Eval0} & \textbf{Eval1} & \textbf{Eval2-80} & \textbf{-90} & \textbf{-100} \\
    \midrule

    Qwen3-8B & 0.6 & 0.6 & 0.6 & 0.6 & 0.0 \\

    Qwen3-8B w/ CorrectBench \cite{qiu2024correctbench} & 47.4 & 25.6 & 23.7 & 23.1 & 21.8 \\

    \midrule 

    Qwen3-8B \,\, w/ PRO-V sys & 64.7 & 39.1 & 35.9 & 32.1 & 28.2 \\

    PRO-V-DS-8B w/ PRO-V sys & 93.6 & 50.6 & 35.9 & 34.0 & 27.6 \\

    \textbf{PRO-V-R1-8B w/ PRO-V sys} & \textbf{94.9} & \textbf{57.7} & \textbf{44.2} & \textbf{37.8} & \textbf{34.0} \\

    \bottomrule
  \end{tabularx}
\end{table}
\label{sec:ablation}

We conduct an ablation study to dissect the contribution of each component in our framework, which is presented in Tab.~\ref{tab:ablation_prov_framework}.

We first establish baselines. A vanilla Qwen3-8B model almost completely fails on VerilogEval-v2. 
We then evaluate CorrectBench~\cite{qiu2024correctbench}, the SOTA agentic system, which improves the scores (Eval1: $25.6$, Eval2-100: $21.8$). 
However, its performance is constrained by its design. CorrectBench employs a complex hybrid generation strategy where the agent must produce multiple languages (e.g., HDL, Python). This multi-language requirement increases the cognitive load on the LLM, often leading to hallucinations and semantic errors.

In contrast, using the same base LLM within our PRO-V sys already yields superior gains (Eval1: $39.1$, Eval2-100: $28.2$). 
This superior baseline performance is attributed to our design, which leverages a "program-tool-enhanced-reasoning" approach. The agent's task is simplified to only generating Python code, which then use  tool interaction. This single-language focus significantly reduces hallucinations and improves reliability. Moreover, adding post-training on top of the sys (\nickname) dramatically improves the test harness correctness, boosting Eval0 from $64.7$ to $94.9$ and Eval1 to $57.7$, reaching $44.2/37.8/34.0$ on Eval2-80/90/100, respectively.

\begin{figure}[t]
    \centering
    \includegraphics[width=\linewidth]{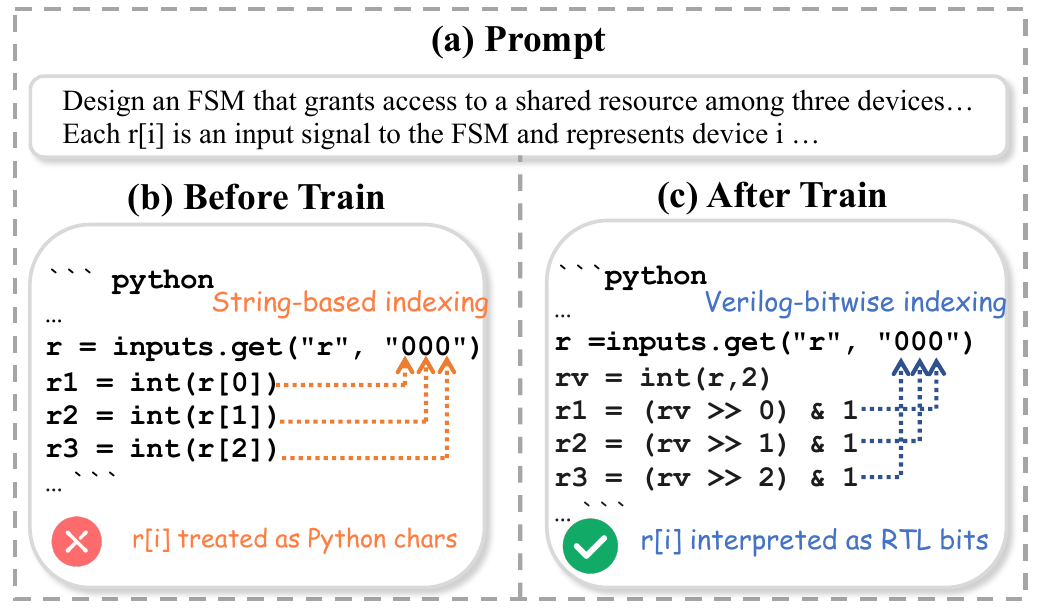}
 \caption{
Post-training case study (task 2013\_q2afsm). 
(a) Input prompt. 
(b) Base LLM fails by treating input as a Python string. 
(c) Our model correctly generates Verilog bitwise operations.
}
 \label{fig:casestudy}
    
\end{figure}

\noindent\textbf{Case Study: Impact of Post-Training}. Fig.~\ref{fig:casestudy} presents a compelling case study on the impact of our post-training process on semantic reasoning.
Given a prompt to design an FSM with a multi-bit input vector \texttt{r} (a), the base LLM (b) misinterprets the vector as a Python string.
This fundamental misunderstanding leads it to generate incorrect, string-based indexing (e.g., \texttt{int(r[0])}) to access what it believes are characters.
In contrast, our post-trained model (c) correctly grasps the hardware context and Verilog semantics, identifying \texttt{r} as an RTL bit vector and generating the proper Verilog-compliant bitwise operations to access individual bits.
This demonstrates that our post-training successfully imbues the model with high-quality reasoning for RTL verification.

\begin{figure}[t]
    \centering
    \includegraphics[width=\linewidth]{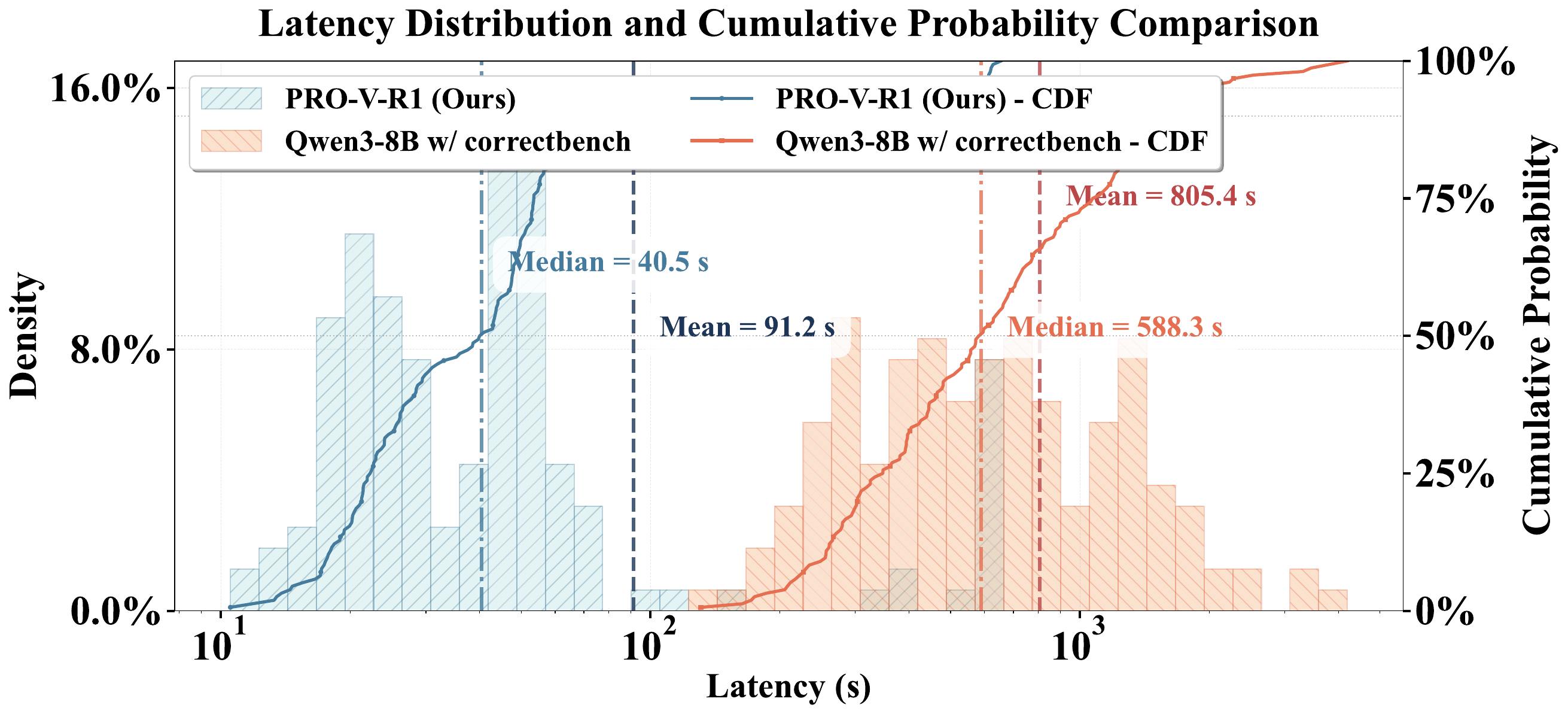}
    \caption{
Latency distribution comparison between \nickname  and the CorrectBench (Qwen3-8B) baseline.
}
\label{fig:efficiency}

\end{figure}

\noindent\textbf{Efficiency Analysis}.
Our framework is also highly efficient. As analyzed in Fig.~\ref{fig:efficiency}, the SOTA agent system, CorrectBench (using Qwen3-8B), has a mean latency of \textbf{805.4 s} per task. In stark contrast, our method completes verification with a mean latency of only \textbf{91.2 s} and a median of \textbf{48.3 s}, realizing an \textbf{8.8$\times$ speedup} in mean latency. This efficiency gain stems from two key factors. First, our training method ensures the model maintains high-quality reasoning traces, which greatly reduces reasoning time. Second, we avoid the primary bottleneck found in CorrectBench---its computationally expensive self-check  mechanism, which requires the LLM to generate 20 different RTL implementations simply to verify its own output. Our method replaces this by generating $N=5$ Python programs in parallel per iteration and typically succeeding within $1$-$2$ self-consistency iterations (max $3$), reducing the required LLM inference steps and overall latency.





\section{Conclusion}
\label{sec:conclusion}

In this paper, we present \nickname, the first trainable open-source agentic verification framework that couples LLMs with program-in-the-loop tools for automated RTL testbench generation. Centered on the PRO-V system, \nickname reduces verification to Python-only functional reference modeling, mitigating semantic mismatches between Verilog and Python. Integrated with SFT on simulation-validated multi-agent trajectories and GRPO-based RL with verification-structured rewards, our 8B PRO-V-R1 model achieves \textbf{57.7\%} functional correctness (Eval1) and \textbf{34.0\%} full-mutant robustness (Eval2-100) on VerilogEval-v2, surpassing GPT-4o (43.0\% / 27.6\%) and the SOTA agentic baseline CorrectBench (25.6\% / 21.8\%), while delivering a \textbf{8.8$\times$} speed up (91s vs.\ 805s per task). \nickname thus represents a concrete step toward practical, reproducible, and efficient LLM-based RTL verification, offering a robust blueprint for future domain-specialized verification agents. 

\begin{acks}
We thank NVIDIA for providing access to and support for cloud services to perform model training.
\end{acks}

\bibliographystyle{ACM-Reference-Format}
\bibliography{sample-base}

\end{document}